\documentclass[conference,letterpaper,10pt]{IEEEtran}
\IEEEoverridecommandlockouts

\usepackage{cite}
\usepackage{amsmath,amssymb,amsfonts}
\usepackage{graphicx}
\usepackage{textcomp}
\usepackage{booktabs}
\usepackage{multirow}
\usepackage{url}
\usepackage[hidelinks,breaklinks=true]{hyperref}

\IEEEpubid{\parbox{\textwidth}{\footnotesize \textcopyright~2026 IEEE. Personal use of this material is permitted. Permission from IEEE must be obtained for all other uses, in any current or future media, including reprinting/republishing this material for advertising or promotional purposes, creating new collective works, for resale or redistribution to servers or lists, or reuse of any copyrighted component of this work in other works.}}

\begin{document}

\title{Spectral Outliers Reveal Dominant Learned Structure in Transformer Attention
\thanks{Code is available at \url{https://github.com/Kasun-Dewage/spectral-outliers-attention}.}}

%
%
\author{
\IEEEauthorblockN{
Kasun Dewage\IEEEauthorrefmark{1},
Marianna Pensky\IEEEauthorrefmark{1},
Suranadi De Silva\IEEEauthorrefmark{1},
and T.~H. Bandara\IEEEauthorrefmark{2}
}

\IEEEauthorblockA{
\IEEEauthorrefmark{1}\textit{University of Central Florida}\\
Orlando, Florida, USA\\
\texttt{\{KasunTharuka.Dewage, Marianna.Pensky, su966204\}@ucf.edu}
}

\IEEEauthorblockA{
\IEEEauthorrefmark{2}\textit{Virtusa}\\
\texttt{thbandara@virtusa.com}
}
}
\maketitle

\begin{abstract}
We apply Marchenko--Pastur (MP) random matrix theory to pre-trained attention weights in order to separate each projection matrix into a random-like bulk and a set of spectral outliers. We validate this decomposition causally: zeroing the MP-identified outliers (signal) in Mistral-7B drives HellaSwag, MMLU, and PIQA close to random-chance performance, whereas zeroing a count-matched subset of bulk singular values causes smaller but non-negligible degradation. Across 11 pre-trained transformers we identify five recurring patterns: spectral outliers encode a dominant component of the learned structure; Q~projections carry the most outliers; V~projections under grouped-query attention lack a clean signal/noise separation; entry-level outliers form structured row-bands in Q and column-bands in O; and specific residual-stream dimensions persist as band outliers across layers in K and O. We close by outlining how these observations could inform parameter-efficient fine-tuning and structured pruning.
\end{abstract}


\section{Introduction}
\label{sec:intro}

Modern pretrained transformers~\cite{vaswani2017attention} contain billions of parameters, yet not all of them carry the same informational content. Mechanistic interpretability seeks to identify which components encode learned behaviors~\cite{elhage2021mathematical,olsson2022context}, while parameter-efficient methods such as LoRA~\cite{hu2022lora} exploit low-rank structure in weight updates but offer limited guidance on \emph{which} projections or layers should receive the largest adaptation budget.

Random matrix theory (RMT) provides a principled framework for this problem. The Marchenko--Pastur law~\cite{marchenko1967distribution} characterizes the eigenvalue distribution of large random covariance matrices: values exceeding the MP upper edge are unlikely under a purely random model and can therefore indicate structured signal. Prior work~\cite{martin2021implicit,yang2023spectral} focuses mainly on aggregate spectral statistics rather than on the spatial distribution of structure, and typically does not causally validate whether the identified structure is functionally important. Martin and Mahoney~\cite{martin2021implicit} showed that state-of-the-art DNNs often exhibit heavy-tailed spectral distributions rather than the clean MP bulk assumed by the spiked model; our use of the MP framework should therefore be read as an approximation that nonetheless yields empirically useful signal/noise decompositions (Section~\ref{sec:limitations}).

We go beyond scalar summaries and introduce: (1)~\textbf{entry-level outlier heatmaps} revealing row--column positions of anomalously large weights across Q, K, V, O; (2)~\textbf{cross-layer residual-stream alignment analysis} identifying residual-stream dimensions persistently acting as band outliers; and (3)~\textbf{targeted ablation experiments} providing causal evidence that the detected structures are functionally important. We analyze 11~pretrained models spanning encoder-only (BERT~\cite{devlin2019bert}, RoBERTa~\cite{liu2019roberta}) and decoder-only (OPT~\cite{zhang2022opt}, LLaMA~1/2/3~\cite{touvron2023llama1,touvron2023llama2,grattafiori2024llama3}, Mistral~\cite{jiang2023mistral}, Qwen~2.5~\cite{qwen2024qwen25}, Phi-3~\cite{abdin2024phi3}) designs.

Our contributions are:
\begin{enumerate}
    \item MP analysis across six architecture families, revealing recurring spectral patterns in attention weights.
    \item Entry-level outlier heatmaps showing a consistent spatial organization: row-bands in Q and column-bands in O.
    \item Evidence that V projections under GQA~\cite{ainslie2023gqa} contain dramatically fewer MP spectral outliers, suggesting a different signal/noise regime for value computation.
    \item Cross-layer alignment analysis revealing persistent residual-stream channels in K and O.
    \item Causal ablation experiments showing that spectral outliers encode a dominant functionally important component of attention weights, with per-component criticality following K $>$ Q $\gg$ V in the tested setting.
\end{enumerate}

\begin{figure*}[t]
\centering
\includegraphics[width=\textwidth]{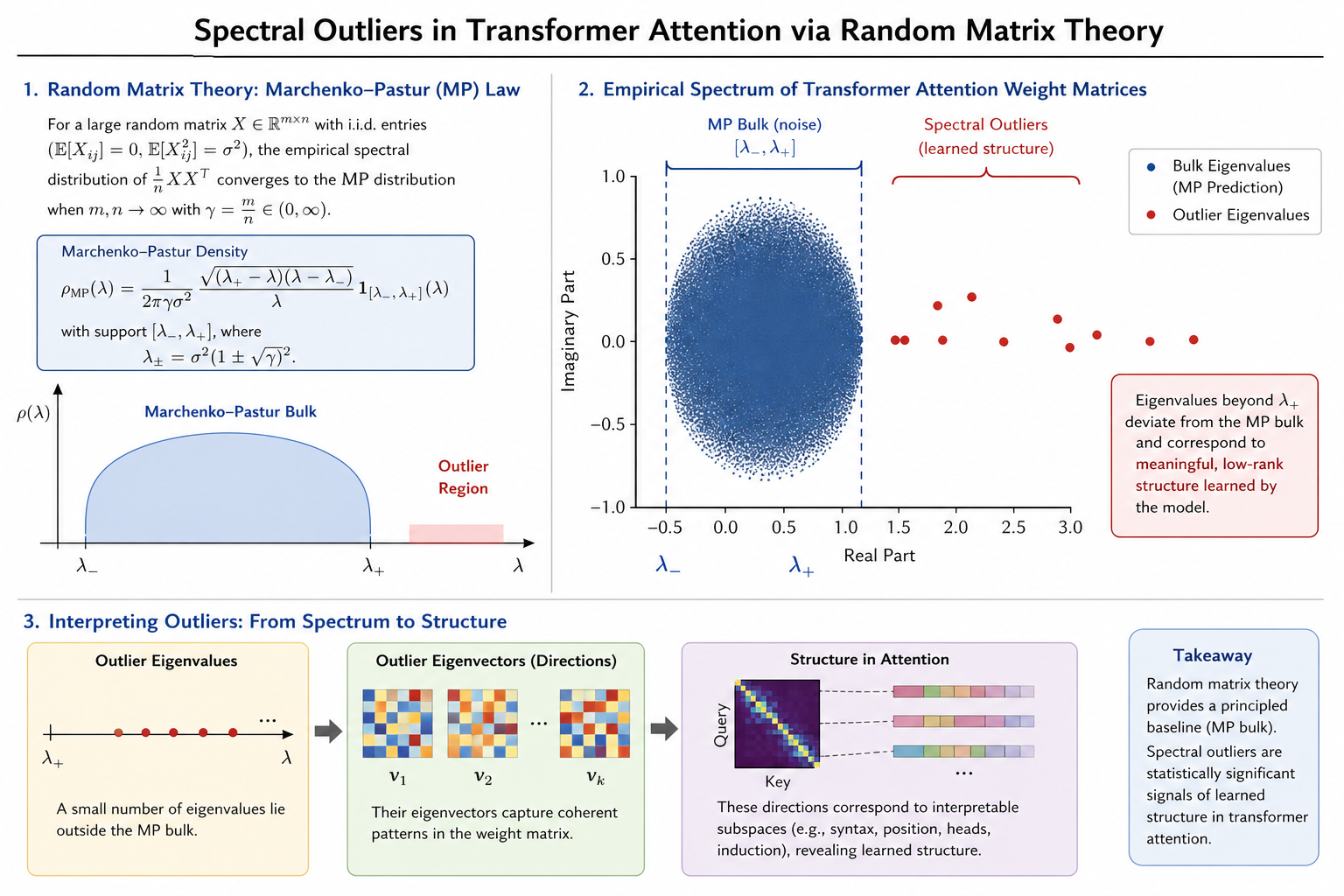}
\caption{Spectral density of transformer weight matrices illustrating the \textit{Marchenko--Pastur} bulk and outlier eigenvalues (signal) corresponding to learned structure.}
\label{fig:mp_spectrum}
\end{figure*}

\section{Method}
\label{sec:method}

\subsection{Marchenko Pastur Spectral Analysis}

Given a weight matrix $W \in \mathbb{R}^{m \times n}$, we compute its singular values $\{s_i\}$ and analyze the nonzero spectrum through squared singular values. The MP threshold is used as a primary separation between a random-like bulk and unusually large spectral components. Let $\gamma = \max(m,n)/\min(m,n)$. We estimate the noise variance as $\hat{\sigma}^2 = \text{median}(s_i^2) / (1 + \gamma)$ and define the MP upper edge as:
\begin{equation}
    \lambda_+ = \hat{\sigma}^2 \left(1 + \sqrt{\gamma}\right)^2.
    \label{eq:mp_threshold}
\end{equation}
Singular values with $s_i^2 > \lambda_+$ are classified as \textbf{spectral outliers (signal)}; Fig.~\ref{fig:mp_spectrum} illustrates this decomposition. We report the outlier count and the spectral energy ratio:
\begin{equation}
    R = \frac{\sum_{i: s_i^2 > \lambda_+} s_i^2}{\sum_i s_i^2}.
    \label{eq:energy_ratio}
\end{equation}
This threshold is only approximate, because trained transformer weights are not random Gaussian matrices and may exhibit heavy-tailed spectra. We therefore treat the MP split as a diagnostic signal/noise proxy rather than as proof that all below-threshold components are noise.

\subsection{Entry-Level Outlier Detection}

For each weight matrix, we compute the standard deviation $\sigma_W$ over all entries and flag positions $(i,j)$ where $|W_{i,j}| > 4\sigma_W$. These are visualized as heatmaps, revealing spatial patterns that are invisible to spectral methods alone.

\subsection{Cross-Layer Residual-Stream Alignment}
\label{sec:method_alignment}

For each layer $\ell$ and projection $P \in \{Q, K, V, O\}$ we compute Euclidean norms along the residual-stream axis of the weight matrix: column norms for Q, K and V, and row norms for O. Columns or rows whose norm exceeds $\mu + 4\sigma$ (where $\mu$ and $\sigma$ are the mean and standard deviation of the norms across all columns or rows) are flagged as \textbf{band outliers}. For Q, K, V matrices of shape $(n_h \cdot d_h) \times d$, outlier columns identify input residual-stream dimensions producing anomalously large projections; for O matrices of shape $d \times (n_h \cdot d_h)$, outlier rows identify output residual-stream dimensions receiving concentrated attention contributions.

A dimension is \textbf{persistent} if it appears as a band outlier in $\geq 3$ layers. We further compute the \textbf{cross-component hit count} as the number of Q, K, V, and O components in which the same dimension appears as a band outlier in at least one layer. Dimensions included in the cross-component table are persistent in at least one component.

\subsection{Ablation Methodology}
\label{sec:method_ablation}

Given $W = U \Sigma V^T$, we apply the MP threshold to classify each singular value as outlier (signal) or bulk. Two strategies:
\begin{itemize}
    \item \texttt{spectral\_outliers}: Zero singular values \emph{above} $\lambda_+$ (remove signal, keep below-threshold components).
    \item \texttt{spectral\_random\_bulk}: Zero a randomly selected subset of singular values \emph{below} $\lambda_+$, with subset size $\lceil f \cdot N_{\mathrm{outlier}}\rceil$, where $N_{\mathrm{outlier}}$ is the number of MP-identified outlier singular values in the same matrix. This provides a count-matched diagnostic comparison against outlier removal while keeping the MP outliers intact.
\end{itemize}
For per-component analysis, the \texttt{entry\_outliers} strategy zeros all entries satisfying $|w_{ij}| > 4\sigma_W$. Ablations are applied before evaluation on HellaSwag~\cite{zellers2019hellaswag}, MMLU~\cite{hendrycks2021mmlu}, and PIQA~\cite{bisk2020piqa} under zero-shot conditions using \texttt{lm-evaluation-harness}~\cite{gao2023evalharness}. The \texttt{spectral\_random\_bulk} results are mainly diagnostic comparisons against outlier removal rather than a full statistical characterization of all possible random bulk subsets.

\subsection{Models Analyzed}

Table~\ref{tab:models} summarizes the 11 pretrained models, spanning six architecture families, used throughout this study.

\begin{table}[t]
\centering
\caption{Models analyzed. GQA = Grouped-Query Attention~\cite{ainslie2023gqa}, MHA = Multi-Head Attention~\cite{vaswani2017attention}.}
\label{tab:models}
\footnotesize
\begin{tabular}{@{}lccccl@{}}
\toprule
\textbf{Model} & $d$ & $L$ & $n_h$ & $n_{kv}$ & \textbf{Attn.} \\
\midrule
BERT-base~\cite{devlin2019bert} & 768 & 12 & 12 & 12 & MHA \\
RoBERTa-base~\cite{liu2019roberta} & 768 & 12 & 12 & 12 & MHA \\
OPT-125M~\cite{zhang2022opt} & 768 & 12 & 12 & 12 & MHA \\
LLaMA-1-7B~\cite{touvron2023llama1} & 4096 & 32 & 32 & 32 & MHA \\
LLaMA-2-7B~\cite{touvron2023llama2} & 4096 & 32 & 32 & 32 & MHA \\
LLaMA-3-8B~\cite{grattafiori2024llama3} & 4096 & 32 & 32 & 8 & GQA \\
Mistral-7B~\cite{jiang2023mistral} & 4096 & 32 & 32 & 8 & GQA \\
Qwen2.5-0.5B~\cite{qwen2024qwen25} & 896 & 24 & 14 & 2 & GQA \\
Qwen2.5-1.5B~\cite{qwen2024qwen25} & 1536 & 28 & 12 & 2 & GQA \\
Qwen2.5-7B~\cite{qwen2024qwen25} & 3584 & 28 & 28 & 4 & GQA \\
Phi-3-mini~\cite{abdin2024phi3} & 3072 & 32 & 32 & 32 & MHA \\
\bottomrule
\end{tabular}
\end{table}

\section{Observational Results}
\label{sec:results}

\subsection{Spectral Outlier Counts and Energy Ratios}

Table~\ref{tab:spectral_summary} reports average spectral outliers (signal) per layer and energy ratios. Three broad patterns emerge across the analyzed models:

\textbf{Pattern 1: Q dominates spectrally.} Across most architectures, Q~projections contain the most spectral outliers. In LLaMA-2-7B, Q averages 1518.7 outliers (87.8\% energy) against 1392.3 (79.0\%) for V.

\textbf{Pattern 2: V is least structured under GQA.} V projections consistently contain fewer outliers in GQA models. In LLaMA-3-8B, V averages only 219.1 outliers (43.1\% energy)---a 6$\times$ reduction relative to Q.

\textbf{Pattern 3: Layer-wise decay.} Energy ratios decrease from early to late layers, suggesting that early layers capture more structured, globally shared representations.

\begin{table}[t]
\centering
\caption{Average MP spectral outliers (signal) per layer and mean energy ratio (\%).}
\label{tab:spectral_summary}
\scriptsize
\setlength{\tabcolsep}{3pt}
\begin{tabular}{@{}l|rr|rr|rr|rr@{}}
\toprule
& \multicolumn{2}{c|}{\textbf{Q}} & \multicolumn{2}{c|}{\textbf{K}} & \multicolumn{2}{c|}{\textbf{V}} & \multicolumn{2}{c}{\textbf{O}} \\
\textbf{Model} & avg & e\% & avg & e\% & avg & e\% & avg & e\% \\
\midrule
\multicolumn{9}{l}{\textit{Encoder-Only (MHA)}} \\
\midrule
BERT-base & 275 & 86.5 & 276 & 86.7 & 268 & 82.5 & 261 & 79.7 \\
RoBERTa-base & 282 & 87.5 & 282 & 87.6 & 271 & 83.6 & 253 & 79.6 \\
\midrule
\multicolumn{9}{l}{\textit{Decoder-Only (MHA)}} \\
\midrule
OPT-125M & 299 & 91.8 & 296 & 90.5 & 271 & 83.5 & 267 & 82.7 \\
LLaMA-1-7B & 1509 & 88.7 & 1509 & 88.5 & 1391 & 79.9 & 1398 & 79.7 \\
LLaMA-2-7B & 1519 & 87.8 & 1517 & 87.5 & 1392 & 79.0 & 1393 & 79.1 \\
Phi-3-mini & --- & --- & --- & --- & --- & --- & 1051 & 79.8 \\
\midrule
\multicolumn{9}{l}{\textit{Decoder-Only (GQA)}} \\
\midrule
LLaMA-3-8B & 1535 & 90.1 & 341 & 75.4 & 219 & 43.1 & 1490 & 86.2 \\
Mistral-7B & 1511 & 87.5 & 341 & 74.7 & 212 & 43.6 & 1450 & 84.6 \\
Qwen2.5-0.5B & 350 & 93.6 & 44 & 71.1 & 13 & 15.6 & 349 & 91.9 \\
Qwen2.5-1.5B & 587 & 92.7 & 86 & 71.2 & 27 & 18.2 & 570 & 88.4 \\
Qwen2.5-7B & 1324 & 89.0 & 170 & 68.8 & 73 & 25.7 & 1290 & 85.6 \\
\bottomrule
\end{tabular}
\end{table}

\subsection{The GQA Effect on Value Projections}

The spectral summary in Table~\ref{tab:spectral_summary} shows a dramatic reduction in V-projection structure under GQA: transitioning from MHA (LLaMA-1/2) to GQA (LLaMA-3, Mistral, Qwen) reduces V outlier counts by 6--27$\times$ while Q counts remain high. Under GQA, shared key--value heads appear to operate in a different spectral regime, with representational capacity concentrating in Q and O. This suggests that the reduced number of KV heads produces a regime in which the V signal is spectrally diffuse: learned structure is distributed across many singular values rather than concentrated in a few dominant ones.

\subsection{Entry-Level Outlier Spatial Patterns}

\begin{figure}[t]
\centering
\includegraphics[width=0.48\columnwidth]{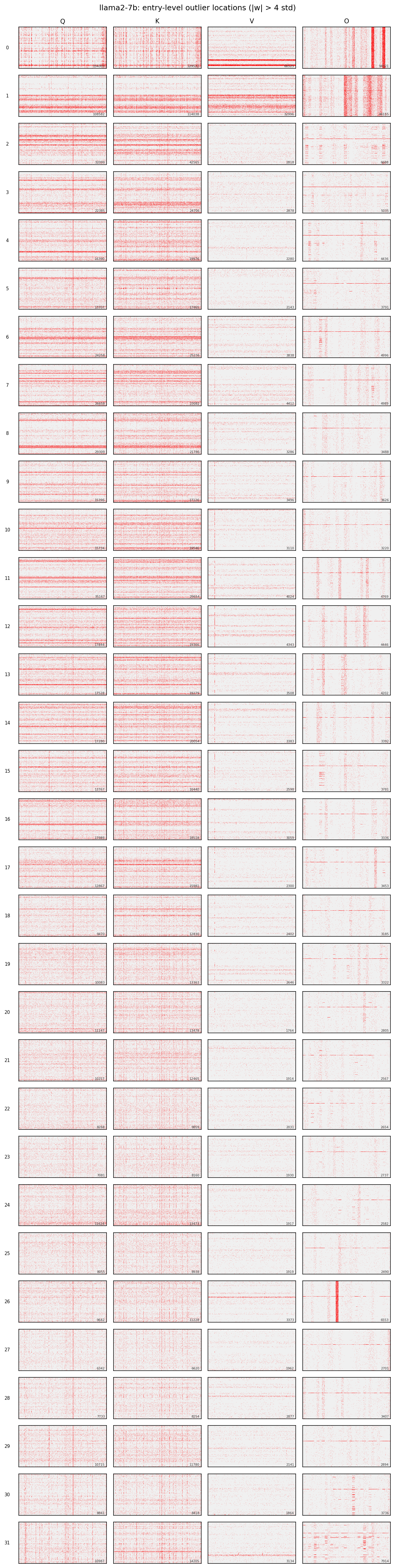}
\includegraphics[width=0.48\columnwidth]{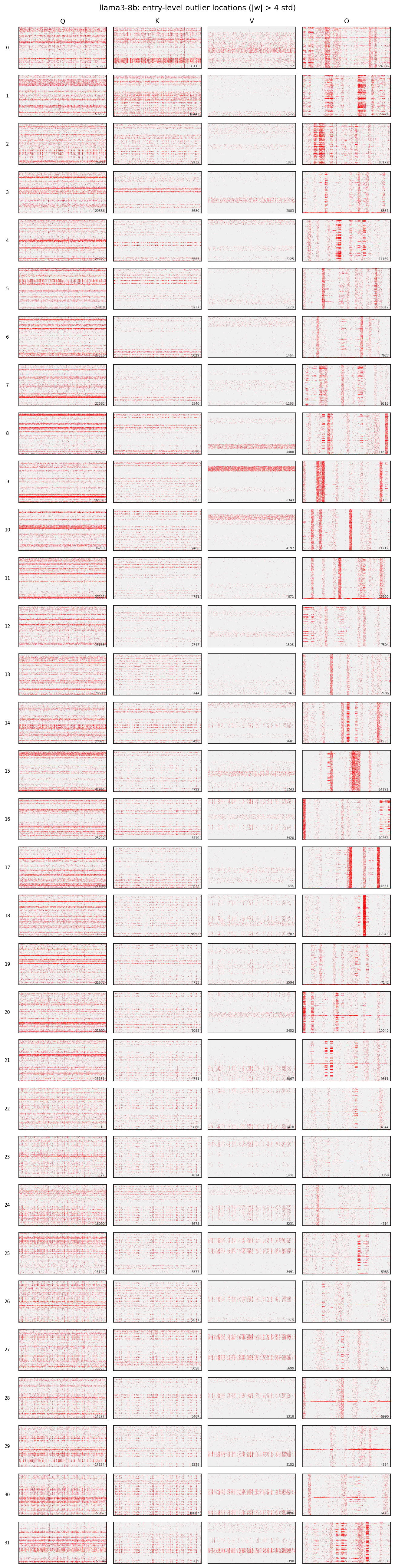}
\caption{Entry-level outlier locations for \textbf{LLaMA-2-7B (MHA, left)} and \textbf{LLaMA-3-8B (GQA, right)}. Under GQA, V becomes extremely sparse while Q retains row-bands and O shows column-bands.}
\label{fig:heatmaps_llama}
\end{figure}

\begin{figure}[t]
\centering
\includegraphics[width=0.48\columnwidth]{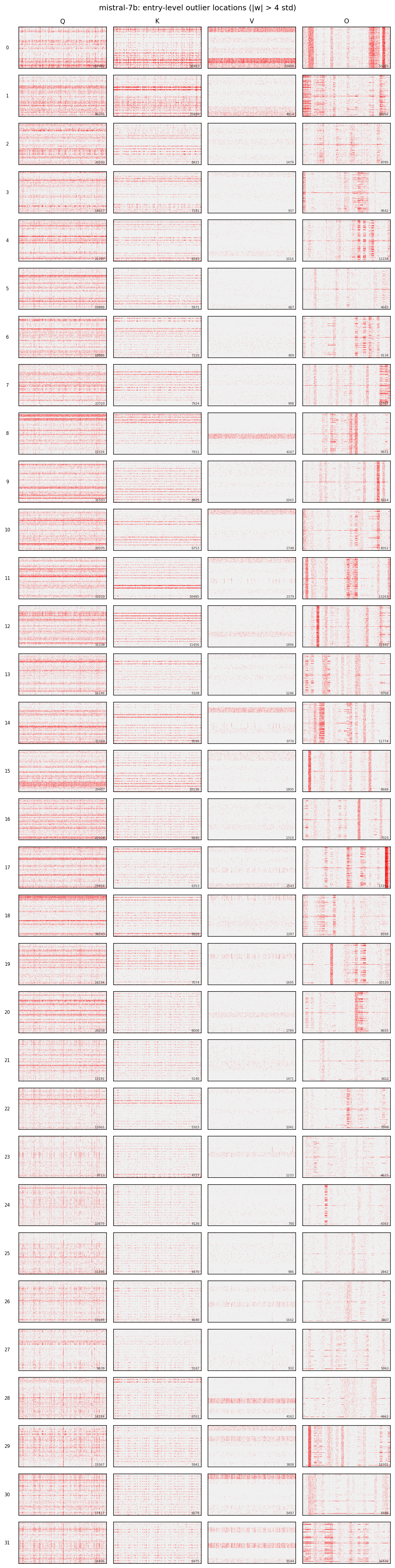}
\includegraphics[width=0.48\columnwidth]{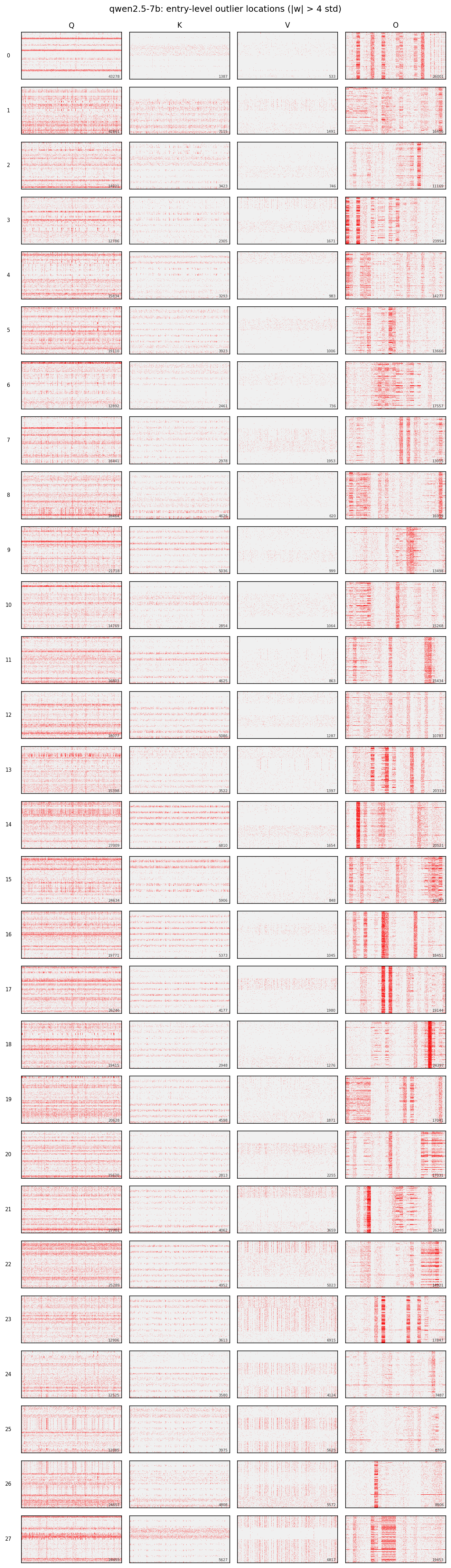}
\caption{Mistral-7B (left) and Qwen2.5-7B (right). Both GQA models reproduce the Q-dense / V-sparse / O-banded pattern.}
\label{fig:heatmaps_additional}
\end{figure}

The heatmaps (Figs.~\ref{fig:heatmaps_llama}--\ref{fig:heatmaps_additional}) reveal coherent spatial structures: \textbf{row-bands in Q} (corresponding to specific attention heads with disproportionate learned structure), \textbf{column-bands in O} (indicating preferred output dimensions for attention contributions), and \textbf{sparse V under GQA} (consistent with low spectral outlier counts). Table~\ref{tab:entry_outliers} quantifies these counts: under GQA, Q and O together account for roughly 85--93\% of all entry-level outliers, against a near-uniform 25\% per projection under MHA. Outlier density in O peaks in middle layers (8--20), mirroring known importance of middle-layer representations~\cite{jawahar2019bert}. We emphasize spatially repeated bands rather than individual high-magnitude entries, since isolated $4\sigma$ entries can occur by chance in very large matrices.

\begin{table}[t]
\centering
\caption{Total entry-level outlier counts ($|W_{ij}|>4\sigma_W$) summed over all layers, with the percentage share of each model's total. Qwen2.5-1.5B and Phi-3-mini are omitted.}
\label{tab:entry_outliers}
\scriptsize
\setlength{\tabcolsep}{2.2pt}
\renewcommand{\arraystretch}{0.92}
\resizebox{\columnwidth}{!}{%
\begin{tabular}{@{}lrrrrrrrr@{}}
\toprule
\textbf{Model} & \textbf{Q} & \textbf{Q\%} & \textbf{K} & \textbf{K\%} & \textbf{V} & \textbf{V\%} & \textbf{O} & \textbf{O\%} \\
\midrule
\multicolumn{9}{l}{\textit{MHA models}} \\
\midrule
BERT-base    & 3{,}305  & 25.5 & 3{,}306  & 25.5 & 3{,}213  & 24.8 & 3{,}127  & 24.1 \\
RoBERTa-base & 3{,}381  & 25.9 & 3{,}388  & 25.9 & 3{,}251  & 24.9 & 3{,}037  & 23.3 \\
OPT-125M     & 3{,}593  & 26.4 & 3{,}547  & 26.1 & 3{,}257  & 23.9 & 3{,}209  & 23.6 \\
LLaMA-1-7B   & 48{,}295 & 26.0 & 48{,}288 & 26.0 & 44{,}521 & 24.0 & 44{,}733 & 24.1 \\
LLaMA-2-7B   & 48{,}599 & 26.1 & 48{,}542 & 26.1 & 44{,}553 & 23.9 & 44{,}583 & 23.9 \\
\midrule
\multicolumn{9}{l}{\textit{GQA models}} \\
\midrule
LLaMA-3-8B   & 49{,}133 & 42.8 & 10{,}921 & 9.5 & 7{,}012 & 6.1 & 47{,}687 & 41.6 \\
Mistral-7B   & 48{,}365 & 43.0 & 10{,}915 & 9.7 & 6{,}785 & 6.0 & 46{,}406 & 41.3 \\
Qwen2.5-0.5B & 8{,}400  & 46.3 & 1{,}059  & 5.8 & 315     & 1.7 & 8{,}366  & 46.1 \\
Qwen2.5-7B   & 37{,}080 & 46.4 & 4{,}758  & 5.9 & 2{,}046 & 2.6 & 36{,}107 & 45.1 \\
\bottomrule
\end{tabular}%
}
\end{table}

\subsection{Cross-Layer Residual-Stream Alignment}
\label{sec:results_alignment}

Table~\ref{tab:persistent_dims} reports persistent dimensions per component. K dominates cross-layer persistence in MHA models (BERT: K=7 vs Q=3; LLaMA-2: K=52 vs Q=33), while V has near-zero persistence under MHA. GQA creates V persistence (LLaMA-3: 44 vs LLaMA-2: 1), as fewer KV heads may force stronger specialization.

Specific O dimensions persist across nearly all layers with remarkable consistency (e.g., LLaMA-1 dim~3840: 32/32 layers; LLaMA-2 dim~1512: 32/32 layers; RoBERTa dim~588: 12/12 layers), representing candidate ``write channels.''

\textbf{Pattern 4: Persistent residual-stream channels.} A small set of residual-stream dimensions acts as a band outlier in K and O across nearly all layers, forming candidate privileged communication channels.

\begin{table}[t]
\centering
\caption{Persistent residual-stream dimensions ($\geq 3$ layers as a band outlier) per component. Mistral-7B, the Qwen2.5 family and Phi-3-mini are omitted.}
\label{tab:persistent_dims}
\footnotesize
\setlength{\tabcolsep}{4pt}
\begin{tabular}{@{}lrrrrc@{}}
\toprule
\textbf{Model} & \textbf{Q} & \textbf{K} & \textbf{V} & \textbf{O} & \textbf{Union} \\
\midrule
BERT-base & 3 & 7 & 0 & 2 & 11 \\
RoBERTa-base & 8 & 10 & 0 & 1 & 11 \\
OPT-125M & 15 & 17 & 0 & 12 & 24 \\
\midrule
LLaMA-1-7B & 21 & 51 & 2 & 15 & 52 \\
LLaMA-2-7B & 33 & 52 & 1 & 17 & 54 \\
LLaMA-3-8B & 87 & 98 & 44 & 43 & 164 \\
\bottomrule
\end{tabular}
\end{table}

Table~\ref{tab:cross_component} shows top cross-component persistent dimensions. In LLaMA-2-7B, dimension~2533 appears in all four components and is persistent in Q, K, and O, while appearing in V in one layer. It appears as a Q outlier in all 32 layers---a simultaneously dominant query direction, frequent key direction, and preferred output channel. Dimension~1512, by contrast, is almost exclusively an O-channel (32/32 layers), serving as a dedicated write-heavy channel.

\begin{table}[t]
\centering
\caption{Top cross-component persistent dimensions. ``Hits'' is the number of Q, K, V, and O components in which the dimension appears at least once; included dimensions are persistent in at least one component.}
\label{tab:cross_component}
\footnotesize
\setlength{\tabcolsep}{3pt}
\begin{tabular}{@{}lcrrrrrr@{}}
\toprule
\textbf{Model} & \textbf{Dim} & \textbf{Hits} & \textbf{Total} & \textbf{Q} & \textbf{K} & \textbf{V} & \textbf{O} \\
\midrule
\multirow{2}{*}{BERT} & 381 & 3 & 16 & 7 & 8 & 0 & 1 \\
 & 308 & 1 & 8 & 0 & 0 & 0 & 8 \\
\midrule
\multirow{2}{*}{OPT} & 174 & 3 & 23 & 10 & 10 & 0 & 3 \\
 & 638 & 1 & 7 & 0 & 0 & 0 & 7 \\
\midrule
\multirow{3}{*}{LLaMA-2} & 2533 & 4 & 58 & 32 & 12 & 1 & 13 \\
 & 3431 & 3 & 61 & 25 & 28 & 0 & 8 \\
 & 1512 & 3 & 37 & 3 & 2 & 0 & 32 \\
\midrule
\multirow{3}{*}{LLaMA-3} & 2977 & 3 & 54 & 24 & 22 & 0 & 8 \\
 & 4055 & 1 & 30 & 0 & 0 & 0 & 30 \\
 & 373 & 3 & 48 & 16 & 23 & 0 & 9 \\
\bottomrule
\end{tabular}
\end{table}

\subsection{3D Attention Stack Visualizations}

\begin{figure}[t]
\centering
\includegraphics[width=0.95\columnwidth]{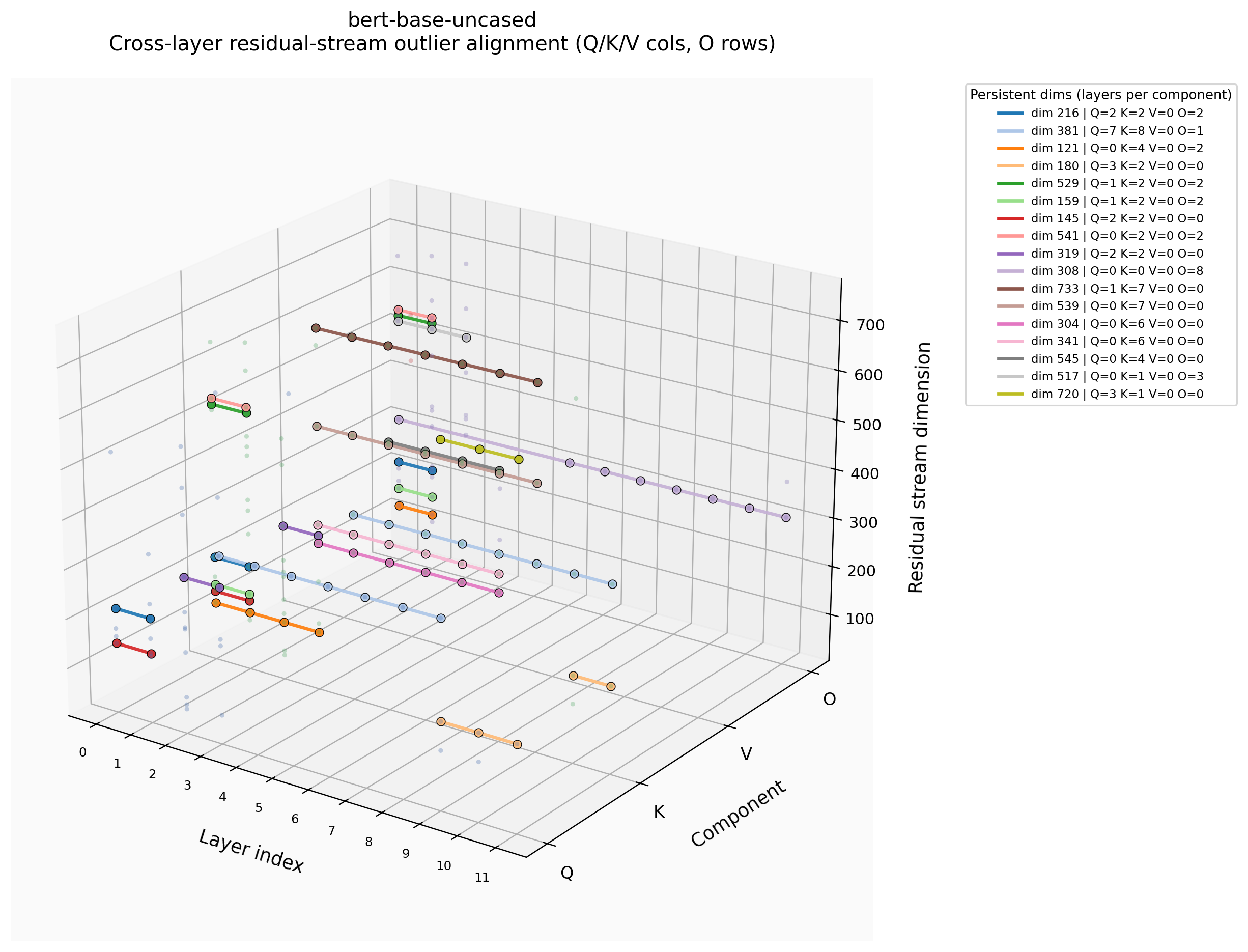}
\caption{3D attention stack for BERT-base. Persistent dimensions concentrate in K and O; V shows none.}
\label{fig:3d_bert}
\end{figure}

\begin{figure}[t]
\centering
\includegraphics[width=0.95\columnwidth]{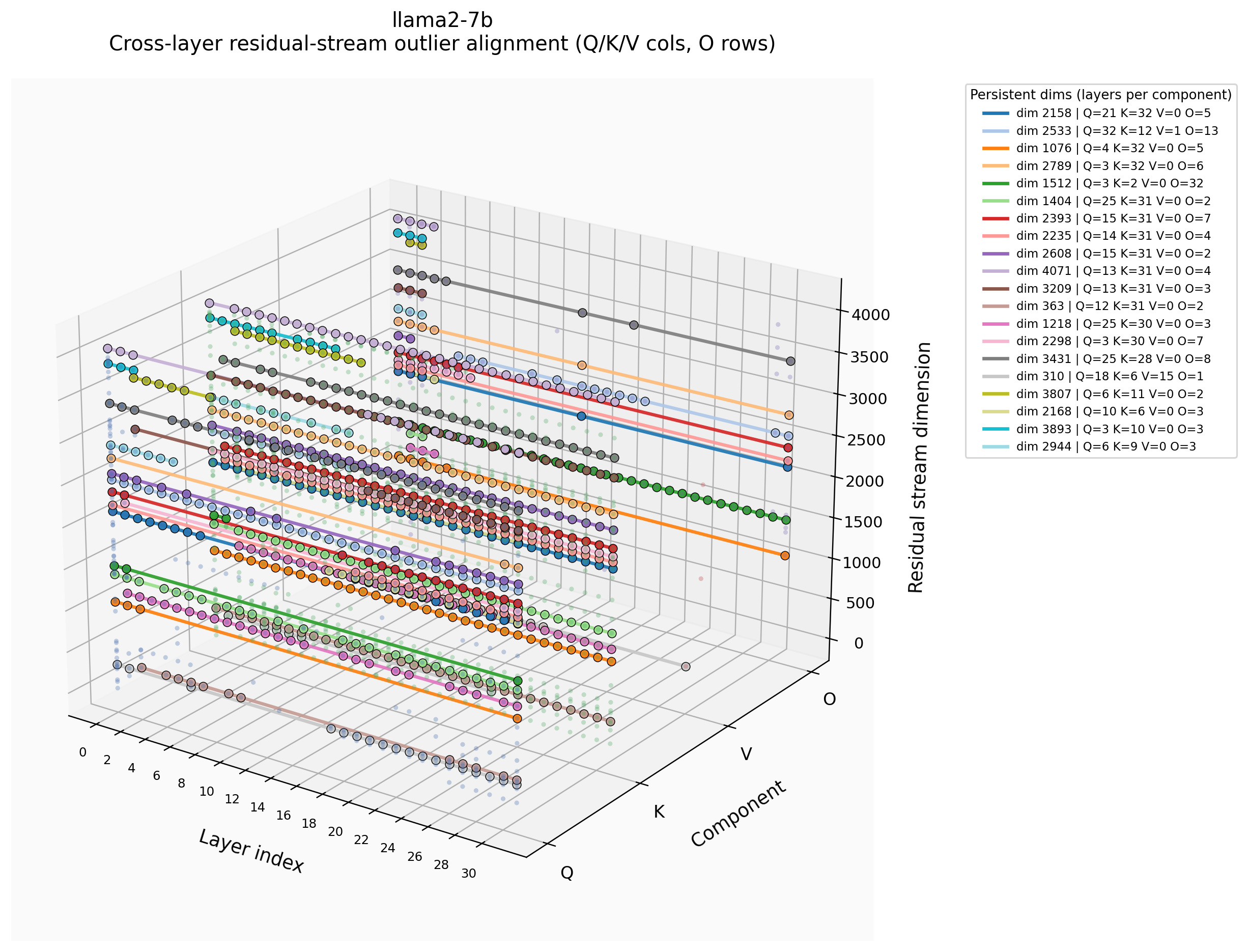}
\caption{3D attention stack for LLaMA-2-7B. Dense persistent dimensions span nearly all 32 layers.}
\label{fig:3d_llama2}
\end{figure}

Figs.~\ref{fig:3d_bert} and~\ref{fig:3d_llama2} present 3D visualizations of cross-layer alignment for BERT-base and LLaMA-2-7B. BERT's stack is sparse with persistence concentrated in K and O spanning only a fraction of layers. LLaMA-2's stack is dense, with dozens of dimensions forming continuous lines across all 32 layers, reflecting deeper architecture and greater specialization.

\section{Causal Validation via Ablation}
\label{sec:ablation}

\subsection{Experimental Design}

We conduct ablations on LLaMA-1-7B~\cite{touvron2023llama1}, LLaMA-2-7B~\cite{touvron2023llama2} (MHA), LLaMA-3-8B~\cite{grattafiori2024llama3}, and Mistral-7B~\cite{jiang2023mistral} (GQA). Table~\ref{tab:ablation_main} reports all configurations.

\begin{table*}[t]
\centering
\caption{Ablation results. $N_\text{zero}$ = singular values/entries zeroed. HS = HellaSwag, PIQA = acc\_norm. Random chance: HS $\approx$ 0.250, MMLU = 0.250, PIQA $\approx$ 0.500.}
\label{tab:ablation_main}
\footnotesize
\setlength{\tabcolsep}{4pt}
\begin{tabular}{@{}llllrrrrrrr@{}}
\toprule
\textbf{Model} & \textbf{Strategy} & \textbf{Comp.} & $f$ & $N_\text{zero}$ & \textbf{HS} & $\Delta$\textbf{HS} & \textbf{MMLU} & $\Delta$\textbf{MMLU} & \textbf{PIQA} & $\Delta$\textbf{PIQA} \\
\midrule
\multicolumn{11}{l}{\textit{Baselines}} \\
\midrule
LLaMA-1-7B & --- & --- & --- & --- & .761 & --- & .352 & --- & .793 & --- \\
LLaMA-2-7B & --- & --- & --- & --- & .760 & --- & .458 & --- & .790 & --- \\
LLaMA-3-8B & --- & --- & --- & --- & .821 & --- & .660 & --- & .812 & --- \\
Mistral-7B & --- & --- & --- & --- & .814 & --- & .627 & --- & .823 & --- \\
\midrule
\multicolumn{11}{l}{\textit{Bulk (below-threshold) removal}} \\
\midrule
LLaMA-1-7B & random\_bulk & QKVO & 0.75 & 139{,}423 & .754 & $-.007$ & .292 & $-.060$ & .791 & $-.002$ \\
LLaMA-2-7B & random\_bulk & QKVO & 0.75 & 139{,}754 & .737 & $-.023$ & .358 & $-.100$ & .773 & $-.017$ \\
LLaMA-2-7B & random\_bulk & QKVO & 1.00 & 186{,}273 & .710 & $-.050$ & .329 & $-.129$ & .752 & $-.038$ \\
\midrule
\multicolumn{11}{l}{\textit{Outlier (signal) removal}} \\
\midrule
Mistral-7B & outliers & QKVO & 1.00 & 112{,}458 & .256 & $-.558$ & .269 & $-.358$ & .508 & $-.315$ \\
\midrule
\multicolumn{11}{l}{\textit{Per-component ablations (LLaMA-3-8B)}} \\
\midrule
LLaMA-3-8B & entry\_outliers & Q & 1.00 & 883{,}337 & .757 & $-.064$ & .488 & $-.172$ & .792 & $-.020$ \\
LLaMA-3-8B & entry\_outliers & K & 1.00 & 234{,}239 & .728 & $-.093$ & .442 & $-.218$ & .770 & $-.042$ \\
LLaMA-3-8B & entry\_outliers & V & 1.00 & 98{,}866 & .793 & $-.028$ & .612 & $-.048$ & .799 & $-.013$ \\
LLaMA-3-8B & random\_bulk & O & 1.00 & 47{,}676 & .789 & $-.032$ & .606 & $-.054$ & .797 & $-.015$ \\
LLaMA-3-8B & random\_bulk & V & 1.00 & 7{,}013 & .649 & $-.172$ & .254 & $-.406$ & .779 & $-.033$ \\
\bottomrule
\end{tabular}
\end{table*}

\subsection{Spectral Outliers Carry a Dominant Learned Structure}

Zeroing all 112{,}458 spectral outliers in Mistral-7B reduces HellaSwag from 0.814 to 0.256, MMLU from 0.627 to 0.269, and PIQA from 0.823 to 0.508, as reported in Table~\ref{tab:ablation_main}. These values are close to random-chance performance, providing causal evidence that MP-identified spectral outliers encode a dominant learned component in this model.

\textbf{Pattern 5: Spectral outliers capture a dominant learned structure.}

Conversely, zeroing count-matched below-threshold singular values preserves much of the model's capability: in LLaMA-1-7B, zeroing 139{,}423 randomly selected below-threshold singular values matched to 75\% of the MP-outlier count causes only a 0.7-point HellaSwag drop. However, MMLU shows greater sensitivity (6--13 point drops), suggesting knowledge-intensive tasks rely on information distributed more broadly across the spectrum, including within components classified as ``bulk'' by the MP threshold. This indicates that while the MP decomposition captures the primary signal, the bulk is not purely noise---it contains secondary learned structure that contributes to knowledge-intensive tasks.

\subsection{Per-Component Criticality and V Bulk Catastrophe}

K has highest per-parameter criticality in the LLaMA-3-8B entry-outlier ablations: zeroing 234{,}239 K entry outliers produces a 21.8-point MMLU drop, while zeroing 883{,}337 Q entry outliers (3.8$\times$ more) produces a 17.2-point drop. V entry outliers are least critical (4.8-point MMLU drop).

The V bulk ablation reveals a regime difference under GQA: zeroing only 7{,}013 below-threshold V singular values causes catastrophic MMLU damage (to 0.254). Table~\ref{tab:mmlu_subcats} breaks these results down by MMLU subcategory. The degradation is broadly uniform across subcategories for the two collapse cases (Mistral outlier removal and LLaMA-3 V bulk removal), whereas the milder per-component ablations preserve social-science and ``other'' accuracy noticeably better than STEM and humanities.

\begin{table}[t]
\centering
\caption{MMLU subcategory accuracy under selected ablations.}
\label{tab:mmlu_subcats}
\footnotesize
\setlength{\tabcolsep}{3pt}
\begin{tabular}{@{}lrrrr@{}}
\toprule
\textbf{Experiment} & \textbf{STEM} & \textbf{Hum} & \textbf{SocSci} & \textbf{Other} \\
\midrule
Mistral outliers QKVO & .286 & .242 & .311 & .251 \\
LL2 random-bulk QKVO $f$=0.75 & .305 & .335 & .386 & .420 \\
\midrule
LL3 entry K & .394 & .375 & .541 & .494 \\
LL3 entry Q & .439 & .403 & .599 & .558 \\
LL3 entry V & .536 & .538 & .717 & .698 \\
LL3 random-bulk V & .252 & .246 & .255 & .265 \\
\bottomrule
\end{tabular}
\end{table}

\section{Connections to Mechanistic Interpretability}
\label{sec:interp}

Our analyses provide three complementary lenses. \textbf{Entry-level heatmaps} connect to the circuits framework~\cite{elhage2021mathematical}: row-bands in Q indicate head-specific structure potentially corresponding to specialized attention behavior, while column-bands in O suggest preferential writing to specific residual stream dimensions~\cite{elhage2022superposition}. \textbf{Cross-layer alignment} reveals that residual-stream communication is not uniform but concentrated on privileged highway dimensions---K-persistent dimensions may encode features consistently attended to, while O-persistent dimensions carry features repeatedly updated by attention layers. \textbf{Causal ablations} close the loop by confirming that these structures are not merely visual artifacts: the K $>$ Q $\gg$ V hierarchy reflects K's sparse but persistent outlier dimensions concentrating disproportionate functional importance in the tested LLaMA-3 setting.

\section{Related Work}
\label{sec:related}

\textbf{RMT for neural networks.} Martin and Mahoney~\cite{martin2021implicit} applied RMT to study implicit regularization via heavy-tailed spectral analysis, demonstrating that well-trained DNNs exhibit heavy-tailed eigenvalue distributions that go beyond the clean MP bulk-plus-outlier picture. Our work operates within the simpler MP framework~\cite{marchenko1967distribution} as a first-order approximation, but validates its utility through causal ablations. Yang \emph{et al.}~\cite{yang2023spectral} analyzed spectral norm scaling conditions for feature learning at large width. We extend these with spatial outlier analysis, cross-layer alignment, and causal validation.

\textbf{Mechanistic interpretability.} The circuits framework~\cite{elhage2021mathematical} and induction head analysis~\cite{olsson2022context} identify functional components via activations. Our weight-space analysis complements these by revealing persistent residual-stream communication channels, validated by ablation.

\textbf{Efficient adaptation and pruning.} LoRA~\cite{hu2022lora} exploits low-rank structure; our MP analysis provides a possible basis for heterogeneous rank allocation---adaptation may benefit from targeting outlier-rich subspaces, with K projections receiving proportionally more capacity despite their smaller dimensions under GQA~\cite{ainslie2023gqa}. The V-projection exception cautions against uniform singular-value thresholding across projections.

\section{Limitations}
\label{sec:limitations}

\textbf{MP framework assumptions.} Our framework uses the standard MP/spiked-model threshold as a first-order approximation, but Martin and Mahoney~\cite{martin2021implicit} showed that trained DNN weight matrices often exhibit heavy-tailed spectral distributions. This means the ``bulk'' identified by the MP threshold is not purely random noise---it may contain structured information that is spectrally diffuse. The MMLU sensitivity to bulk removal (6--13 point drops) empirically confirms this. A heavy-tailed RMT framework could yield a more refined decomposition.

\textbf{Causal claims.} Our ablation demonstrates that MP-identified outliers are functionally important and that removing them can collapse model performance. However, it does not prove that no learned structure exists in the bulk---the bulk removal experiments show modest but non-trivial degradation, especially on knowledge-intensive benchmarks. We therefore characterize outliers as encoding a ``dominant'' rather than exclusive learned structure.

\textbf{Random bulk ablations.} The random-bulk experiments are intended as diagnostic contrasts against MP outlier removal. A more complete robustness study would average random bulk subsets over multiple seeds and report variance.

\textbf{Entry-level thresholding.} The $|W_{ij}|>4\sigma_W$ threshold identifies interpretable spatial bands, but isolated extreme values can occur in large random matrices. Future work should include shuffled-weight and randomly initialized controls to separate true learned spatial organization from threshold artifacts.

\textbf{Noise variance estimation.} The median-based estimator $\hat{\sigma}^2 = \text{median}(s_i^2)/(1+\gamma)$ is approximate---the exact median of the MP distribution depends on $\gamma$ and differs slightly from this formula. This could affect threshold placement for matrices with unusual aspect ratios.

\textbf{Model coverage.} For Phi-3-mini, only O-projection data is reported, because the fused QKV projection prevented separate Q, K, and V extraction. For the same reason, and to respect the page limit, Tables~\ref{tab:entry_outliers} and~\ref{tab:persistent_dims} report a subset of the models in Table~\ref{tab:models}. Patterns observed in models with excluded projection types should therefore be interpreted cautiously.

\section{Conclusion}
\label{sec:conclusion}

We presented a systematic MP analysis of attention weights across 11 transformers, augmented with entry-level outlier heatmaps, cross-layer residual-stream alignment, and causal ablation experiments. We identified five recurring patterns: Q dominance in spectral structure; V sparsification under GQA; structured spatial organization in Q and O; persistent residual-stream highways in K and O; and the causal finding that spectral outliers encode a dominant learned structure, while below-threshold singular values are typically less critical though not purely noise. These findings connect random matrix theory with mechanistic interpretability and offer practical guidance for pruning, compression, and low-rank adaptation. The observed cross-layer structure is complementary to CRAFT~\cite{dewage2026craft}, which exploits correlations across attention layers through a frozen Tucker decomposition for parameter-efficient fine-tuning. Future directions---in particular spectrum-aware LoRA, MP-guided structured pruning, and per-projection low-rank compression---represent concrete pathways for translating these observations into practical efficiency gains for large language models.

\end{document}